\documentclass[11pt]{article}

\usepackage{fullpage}
\usepackage[ruled, linesnumbered, vlined, commentsnumbered]{algorithm2e}
\usepackage{graphicx,subfigure} 
\usepackage{amsfonts,amssymb,amsmath,amsthm,amsopn}	
\usepackage{hhline,booktabs,diagbox,colortbl,multirow,tabularx,threeparttable} 
\usepackage[listings,skins,breakable]{tcolorbox}

\usepackage{enumerate}
\usepackage{authblk}
\usepackage{footnote}
\usepackage{hyperref}
\usepackage{prettyref}
\usepackage{cite}
\usepackage{setspace}
\usepackage{color}
\usepackage{xcolor}  
\usepackage{scrlayer-scrpage}
\usepackage{lscape}
\usepackage{geometry}
\usepackage{framed}
\usepackage{url}

\usepackage{tikz}
\usepackage{pgfplots}
\usetikzlibrary{positioning,shapes,shadows,arrows,calc}
\tikzstyle{component}=[rectangle, draw=black, rounded corners, fill=blue!40, drop shadow, text centered, anchor=north, text=white, minimum height=1cm]
\tikzstyle{arrow}=[->, thick]

\pgfplotsset{compat=1.12}
\usetikzlibrary{intersections}
\usetikzlibrary{pgfplots.statistics}
\usepgfplotslibrary{fillbetween}

\definecolor{myblue}{RGB}{34,31,217}
\definecolor{mycyan}{gray}{.7}
\definecolor{Gray}{gray}{0.9}

\newtheorem{remark}{Remark}

\newcommand{\pref}{\prettyref}

\newrefformat{fig}{Fig.~\ref{#1}}
\newrefformat{tab}{Table~\ref{#1}}
\newrefformat{sec}{Section~\ref{#1}}
\newrefformat{app}{Appendix~\ref{#1}}
\newrefformat{alg}{Algorithm~\ref{#1}}
\newrefformat{property}{Property~\ref{#1}}
\newrefformat{theorem}{Theorem~\ref{#1}}
\newrefformat{corollary}{Corollary~\ref{#1}}
\newrefformat{proposition}{Proposition~\ref{#1}}
\newrefformat{def}{Definition~\ref{#1}}
\newrefformat{eq}{equation~(\ref{#1})}

\title{\vspace{-1ex}\LARGE\textbf{Solving Expensive Optimization Problems in Dynamic Environments with Meta-learning}\footnote{This work has been accepted for publication in \textit{IEEE Transactions on Cybernetics}.}}

\author[1]{\normalsize Huan Zhang}
\author[1]{\normalsize Jinliang Ding}
\author[2]{\normalsize Liang Feng}
\author[3]{\normalsize Kay Chen Tan}
\author[4]{\normalsize Ke Li}
\affil[1]{\normalsize State Key Laboratory of Synthetical Automation for Process Industries, Northeastern University, Shenyang, 110819, China}
\affil[2]{\normalsize College of Computer Science, Chongqing University, Chongqing 400044, China}
\affil[3]{\normalsize Department of Data Science and Artificial Intelligence, Hong Kong Polytechnic University, Hong Kong SAR}
\affil[4]{\normalsize Department of Computer Science, University of Exeter, EX4 4QF, Exeter, UK}
\affil[$\ast$]{\normalsize Email: \texttt{huanzhang0320@163.com}}

\date{}

\begin{document}
\maketitle

\textbf{Abstract:} Dynamic environments pose great challenges for expensive optimization problems, as the objective functions of these problems change over time and thus require remarkable computational resources to track the optimal solutions. Although data-driven evolutionary optimization and Bayesian optimization (BO) approaches have shown promise in solving expensive optimization problems in static environments,  the attempts to develop such approaches in dynamic environments remain rarely explored. In this paper, we propose a simple yet effective meta-learning-based optimization framework for solving expensive dynamic optimization problems. This framework is flexible, allowing any off-the-shelf continuously differentiable surrogate model to be used in a plug-in manner, either in data-driven evolutionary optimization or BO approaches. In particular, the framework consists of two unique components: 1) the meta-learning component, in which a gradient-based meta-learning approach is adopted to learn experience (effective model parameters) across different dynamics along the optimization process. 2) the adaptation component, where the learned experience (model parameters) is used as the initial parameters for fast adaptation in the dynamic environment based on few shot samples. By doing so, the optimization process is able to quickly initiate the search in a new environment within a strictly restricted computational budget. Experiments demonstrate the effectiveness of the proposed algorithm framework compared to several state-of-the-art algorithms on common benchmark test problems under different dynamic characteristics. 



\section{Introduction}
\label{sec:introduction}

Real-world expensive optimization problems are challenging and complex. This is not only due to their non-convex and multi-objective nature but also being susceptible to a range of environmental uncertainties. Among the uncertainties, dynamic optimization problems (DOPs) emerge as prevalent scenarios where the objective function(s) or constraints change over time. Typical instances of DOPs include dynamic job shop scheduling~\cite{ZhangMNZ23}, path planning~\cite{LuTMY22}, and dynamic economic dispatch problem~\cite{WangZLQW11}.

Evolutionary algorithms (EAs) have gained widespread recognition as highly effective tools for tackling DOPs owing to their population-based nature and inherent self-adaptability~\cite{YazdaniCYBJY21A}. However, the computational cost of the objective function evaluations (FEs) is assumed to be cheap or trivial in most EAs. Consequently, it is common for an EA to spend tens of thousands or more FEs even in a single optimization procedure. Unfortunately, this assumption fails to hold in practical optimization scenarios where FEs are expensive to evaluate, either economically or computationally, which is recognized as expensive optimization. The challenge of solving such problems becomes even more pronounced when the optimization environment undergoes dynamic changes over time. This issue is pervasive across various real-world application domains, such as airfoil design~\cite{GuoLOQZZ23}, trauma systems design~\cite{WangJJ16}, and engine calibration optimization~\cite{YuWZFY23}. 

In the literature, data-driven evolutionary optimization\footnote{In the literature, this concept is interchangeably referred to as `surrogate-assisted EAs' (SAEAs)~\cite{HeZGJ23}.}~\cite{JinWCGM19} and Bayesian optimization (BO)~\cite{ShahriariSWAF16} are two widely adopted techniques for solving expensive optimization problems. Both approaches heavily rely on surrogate modeling, which is responsible for approximating the objective function and predicting the fitness of candidate solutions. By incorporating a surrogate model, the search process for subsequent optimization will be guided. In particular, EAs search the candidate solutions based on the predictions of the surrogate model in the data-driven evolutionary optimization approaches, while in BO, an acquisition function is considered to balance the exploitation and exploration over the surrogate landscape. 

Despite extensive research studies having been conducted on data-driven evolutionary optimization and BO for solving static objective function(s) in continuous domains~\cite{HuzLZ23,GuoLOQZZ23}, it is worth noting that there is a lack of approaches specifically tailored for solving expensive DOPs. This gap can be due to two critical challenges. The first challenge arises from the dynamic nature of environments and the changing problem landscapes. When the environment changes, simply restarting the optimization from scratch poses significant difficulties in tracking the shifting optima over time. The second challenge stems from the limited data availability in the expensive optimization scenario. Due to computational constraints, only a limited number of training samples can be obtained at each time step, along with restricted historical data. Consequently, effectively utilizing the limited data to facilitate the search process in a new environment becomes a great challenge.

Meta-learning~\cite{HospedalesAMS22}, also known as `learning to learn', has been extensively studied in many machine learning tasks, such as neural architecture search~\cite{LianZXLLZHG20} and hyperparameter optimization~\cite{WistubaG21}. By leveraging prior knowledge, meta-learning is able to enhance the adaptability of learning in new tasks. It offers a promising approach to mitigating the challenge of expensive optimization problems in dynamic environments. In particular, meta-learning equips models with the ability to learn domain-specific features from related tasks, thus facilitating the task-solving process of a new task. This concept holds promise for application in dynamic optimization, as there exist potential patterns in DOPs across different environments~\cite{YazdaniCYBJY21A}. Moreover, meta-learning can be executed as few-shot learning, extracting effective features from a limited sample size, which aligns with the expensive optimization scenarios well. To the best of our knowledge, there are no existing studies reported in the literature considering meta-learning to improve the optimization performance of expensive DOPs.

Bearing the above in mind, this paper proposes a novel meta-learning-based optimization framework called MLO, specifically designed to address the challenges posed by expensive dynamic optimization problems. The key contributions of this paper are summarized as follows. 
\begin{itemize}
\item To overcome the challenge of changing problem landscapes caused by the dynamic environments, we employ a meta-learning mechanism to learn effective initial parameters for the surrogate model whenever an environmental change is detected, resulting in a strategic initiation at the outset of the new environment.
\item To address the difficulty of working with limited data brought by a restricted computational budget, our framework is designed under the scenario of few-shot learning problems. Meta-learning empowers us to effectively extract domain-specific features from a limited quantity of samples.
\item The proposed algorithm is high flexible in incorporating existing optimization solvers. Firstly, it can be combined with both evolutionary optimization and BO approaches, resulting in meta-learning-based data-driven evolutionary optimization (referred to as \texttt{MLDDEO}) and meta-learning-based Bayesian optimization (referred to as \texttt{MLBO}), respectively. Secondly, the framework is compatible with any continuously differentiable surrogate model.
\item In the empirical study, the proposed framework generates seven algorithm instances by combining two optimization mechanisms, two surrogate models, and three baseline EAs. Through rigorous empirical studies conducted on benchmark problems with diverse dynamic characteristics, we demonstrate the superiority of the proposed algorithms against state-of-the-art peer algorithms in terms of the effectiveness of the solution quality, the efficiency of the computational cost, and flexibility.
\end{itemize}

The remainder of this paper is organized as follows. \pref{sec:background} provides the necessary background of this study and then presents an overview of the existing related work. \pref{sec:method} gives the technical details of the proposed algorithm. The experimental setup is given in \pref{sec:experiment_Setup} and the experimental results and analyses are provided in \pref{sec:empirical_results}. Lastly, \pref{sec:conclusion} draws the conclusions.


\section{Preliminaries}
\label{sec:background}
This section begins with the problem definition of expensive DOP. Next, we present a literature review of expensive dynamic optimization and introduce the concept of meta-learning.

\subsection{Problem Definition}
\label{sec: problem definition}
According to~\cite{LiCY23}, the mathematical form of expensive DOP considered in this paper is as follows:
\begin{equation}
    \begin{aligned}
        &\mathrm{maximize} \quad f({\mathbf{x},t})\\
        &\mathrm{subject\ to } \,\,\, \mathbf{x}\in\mathrm{\Omega}
    \end{aligned},
\end{equation}
where $\mathbf{x} = (x_1,\cdots,x_n)^T \in\mathrm{\Omega}$ is a decision vector, $\mathrm{\Omega}=\mathrm{\Pi}_{i=1}^n[ l_i,u_i]\in\mathbb{R}^n$ is the decision space where $l_i$ and $u_i$ represent the lower and upper bounds of the $i$-th dimension, respectively. $t\in\mathrm{\{1,\cdots,T\}}$ is a discrete time step and $T>1$ refers to the number of time steps. It is important to note that the fitness function $f$ is computationally intensive and subject to changes over time $t$. As a result, the landscapes of the function, as well as the positions of the local and global optima, may fluctuate dynamically. In this dynamic context, the goal of computationally expensive optimization is to monitor the changing optimum at each time step with a strictly restricted computational budget.  

\subsection{Related Work}
BO and data-driven evolutionary optimization are two extensively adopted techniques for tackling expensive optimization problems. While considerable progress has been made in solving static expensive problems, there have been relatively few studies exploring their applicability in dynamic environments. This subsection provides an overview of related work for solving expensive DOPs, including BO and data-driven evolutionary optimization approaches.

In the field of BO, Morales-Enciso and Branke~\cite{MoralesB15} applied the efficient global optimization~\cite{Jones01} framework to propose eight strategies for handling expensive DOPs. The first four are straightforward and simple strategies while the last four are designed to transfer and reuse previous information within the underlying problem-solving process. Moreover, Richter \emph{et al.}~\cite{RichterSCRL20} developed two practical approaches using BO to optimize expensive DOPs. In particular, one is the window approach which selects the most recent samples to train the surrogate model, and the second is the time-as-covariate approach that incorporates the time into the covariance function to learn the influence of time. In~\cite{ChenL21}, a transfer learning strategy was proposed to empower BO for solving expensive DOPs. It applies a multi-output Gaussian process (MOGP) model that measures the relationship between the historical and current samples. In addition, to alleviate the soaring computational cost, a decay mechanism is designed to discard less irrelevant samples. This relies on the assumption that the samples collected from earlier time steps can be less informative than recent ones. Subsequently, in~\cite{LiCY23}, the authors proposed a hierarchical MOGP surrogate model for knowledge transfer and cost-efficiency optimization. Furthermore, this approach incorporates an adaptive source task selection mechanism and a customized warm-start initialization strategy to further enhance the effectiveness of knowledge transfer. A EA with local search is developed to optimize the acquisition function. 

In the domain of data-driven evolutionary optimization, Luo \emph{et al.}~\cite{LuoYYX19} investigated the performance of different SAEAs in solving data-driven dynamic optimization. This method starts with storing elite solutions obtained from previously optimized environments, which are subsequently clustered to construct surrogate models. Based on this, two variants of particle swarm optimization with memory are developed to address DOPs. In~\cite{LiuLZY20} and~\cite{ZhaoHWJLZ23}, the multipopulation technique was employed to solve expensive dynamic optimization. In particular, Liu \emph{et al.}~\cite{LiuLZY20} designed a surrogate-assisted clustering particle swarm optimizer. The affinity propagation clustering method is first applied to create several clusters. Then, local radial basis function surrogates are built in each cluster to assist the particle swarm optimizer. In response to environmental changes, the best-fitness points in each cluster are incorporated into the new cradle swarm. Zhao \emph{et al.}~\cite{ZhaoHWJLZ23} proposed a SAEA based on multi-population clustering and prediction. Three strategies were employed: clustering and differential prediction for initial population selection, multi-population strategies for sample point generation in the surrogate model, and replacing high and low confidence samples in the Co-kriging model with samples from the previous strategy. Liu \emph{et al.}~\cite{LiuLDYJ23} proposed a surrogate-assisted differential evolution with a knowledge transfer approach to address the expensive incremental optimization problem, a type of expensive dynamic optimization problem. In this study, a surrogate-based hybrid knowledge transfer strategy is designed to reuse the knowledge. Furthermore, a two-level surrogate-assisted evolutionary search is proposed to search for the optimum. In addition, some SAEAs have been developed for expensive dynamic multi-objective optimization problems~\cite{FanLT20, LiuW22, ZhangYJQ23}.

In summary, the primary challenges in solving expensive DOPs can be summarized into two key aspects. One challenge stems from the dynamic changing environments. When the environment changes, simply restarting the optimization process from scratch presents considerable difficulties in tracking the time-varying optima. The other challenge arises from the scarcity of available data. Owing to computational constraints, effectively leveraging the limited data to promote the search process in the new environment becomes a main obstacle. Despite the success obtained by the approaches reviewed above, it is worth noting that most existing methods assume that data collected from earlier time steps is less informative than more recent data. This oversight may lead to the neglect of valuable information due to the presence of potential patterns in a DOP across different environments.

\subsection{Motivation}
Meta-learning, a rapidly growing research field in machine learning, focuses on developing algorithms that enable models to learn how to learn. Meta-learning has been successfully applied in many problems such as robotics~\cite{FinnAL17}, the cold-start problem in collaborative filtering~\cite{VartakTMBL17}, and guiding policies by natural language~\cite{CoReyesGSAADAL19}. Readers interested in meta-learning can refer to~\cite{HospedalesAMS22} for a more comprehensive introduction.

Recently, some studies have attempted to adopt meta-learning methods to ‘warm starting’ the optimization process through acquisition function~\cite{VolppPFDFHD20}, weighted surrogate models ensembles~\cite{FeurerLB18}, or deep kernel~\cite{WistubaG21}. From these, we can see that the meta-learning mechanism holds the potential to effectively mitigate the challenges posed by expensive DOPs. This can be attributed to the following reasons. In DOPs, the problems do not change continuously but evolve over discrete time intervals, remaining stable within each interval~\cite{YazdaniCYBJY21,YazdaniCYBJY21A}. This stability allows valuable information to be collected from previously optimized problems due to the presence of potential relevant patterns across different environments. Furthermore, solving expensive optimization problems is challenging due to the high computational cost of function evaluations~\cite{JinWCGM19,HeZGJ23}. Therefore, in data-driven evolutionary optimization and BO, surrogate models are typically constructed to estimate the fitness function and aid in the evaluation of the true function. These approaches aim to leverage data and surrogates to drive optimization, their performance heavily relies on surrogate quality~\cite{LiZWZ21}. The surrogate quality is determined by both the data and the model parameters. Data with higher quality and of larger amounts are useful for building more accurate surrogates~\cite{JinWCGM19}. However, the data collected is limited due to the high computational costs of solving expensive optimization problems. Regarding model parameters, the parameter space of many surrogate models is inherently complex and challenging to optimize~\cite{YangS20}. Some models have high-dimensional parameter spaces such as neural network (NN)~\cite{FeurerH19}. Additionally, certain models exhibit non-convexity with many local optima in their loss functions, such as the marginal likelihood function of Gaussian Process Regression (GPR)~\cite{RasmussenW06} or the loss function of NN~\cite{YangS20}. 

\begin{figure}[t!]
    \centering
    \includegraphics[width=.32\textwidth]{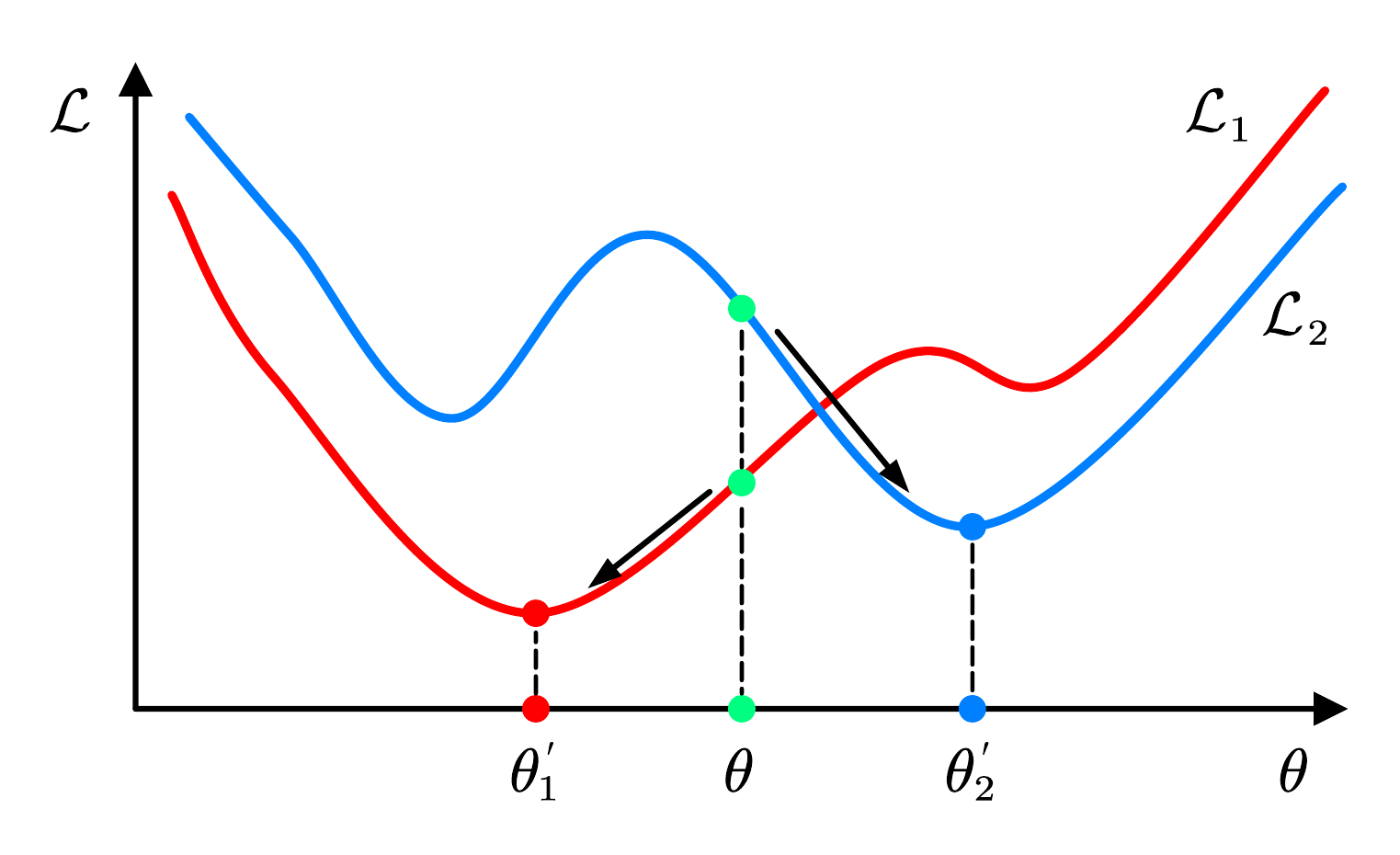}
    \caption{Illustrative example of meta-learning working for two tasks.}
    \label{fig:example}
\end{figure}

The configuration of initial parameters plays a crucial role in the subsequent optimization process~\cite{YuZ20}, especially in complex search spaces. In the traditional data-driven evolutionary optimization and BO approaches, initial parameters are commonly assigned based on rules of thumb or randomly generated using experimental design methods. This approach makes it difficult for the model to track time-varying optima within a limited computational budget. In contrast, our motivation is to utilize meta-learning to learn a favorable initial parameter configuration for the surrogate model, especially when data is limited. By providing effective initial model parameters in complex parameter space, the proposed framework improves the quality of the surrogate model whenever there is an environmental change, thereby aiding in the optimization procedures for solving expensive dynamic optimization problems. Here, we succinctly outline the principles underlying the effectiveness of meta-learning, with an experimental study presented in~\pref{sec: reason for work}. As illustrated in Fig.~\ref{fig:example}, an effective initial parameter $\theta$ is obtained using meta-learning, where `effective' refers to the fact that it is potential for subsequent optimization processes. Specifically, $\mathcal{L}_1$ and $\mathcal{L}_2$ represent the loss functions of two training tasks across a one-dimensional parameter space. We are not concerned with how well $\theta$ performs on the training task, but rather with how well it performs after being trained by meta-learning. $\theta$ can be optimized along the gradient direction for different tasks, making it a promising parameter. With the above in mind, in this paper, we thus propose to leverage the power of meta-learning to address expensive DOPs.

\section{Proposed Algorithm }
\label{sec:method}
In this section, the details of the proposed meta-learning-based approach for solving expensive DOPs are presented. In expensive DOPs, the problem changes over time, thus involving multiple time steps (or environments). The challenge of solving the expensive DOPs is to track the time-varying optimum under a limited computational budget at each time step. Here, we propose to address the expensive DOPs at the current time step by utilizing experience gained from previous time steps. 

In particular, the problem-solving process of an expensive DOP in each new environment is considered a few-shot learning problem~\cite{WangYKN20}. The few-shot learning problem involves multiple related tasks, each with two subsets: a support set for training the model and a query set for evaluating the model's performance. After learning from these related tasks, the goal is to estimate labels or values for a query set when given a support set from a new task. This problem is known as a `few-shot learning problem' due to the limited number of samples used for both training and new tasks. Here, expensive DOPs in different time steps are treated as different tasks. Formally, given an expensive DOP, when the environment changes, previously optimized problems are denoted as related tasks $\{T_{r}\}_{r=0}^{t-1}$. The observations gathered from previous time steps are denoted as sets $\mathcal{D}=\{\mathcal{D}_r\}_{r=0}^{t-1}$. The expensive DOP to be solved in the new environment at time step $t$ is considered the new task $T_{t}$. The observations obtained at the current time step constitute set $\mathcal{D}_{t}$. In what follows, we will present the details of the algorithm framework and the two main ingredients, which are the \texttt{meta-learning} component and the \texttt{adaptation} component.

\subsection{Overall Algorithm Framework}
\label{sec:framework}
In this paper, the proposed framework is highly flexible, enabling us to embed the meta-learning in both evolutionary optimization and BO, which are meta-learning-based data-driven evolutionary optimization (called \texttt{MLDDEO}) and meta-learning-based Bayesian optimization (called \texttt{MLBO}), respectively. In contrast to conventional data-driven evolutionary optimization and BO approaches, the proposed framework leverages meta-learning to efficiently initiate the optimization process when the environment changes. A diagram of the proposed meta-learning-based optimization approach is depicted in Fig.~\ref{fig:flowchart}. The proposed framework comprises two primary components: 1) meta-learning component and 2) adaptation component. The meta-learning part extracts the domain-specific features from previously optimized tasks to generate initial model parameters $\theta^{ml}$ for the surrogate model in the new environment at time step $t$, in both data-driven evolutionary optimization and BO. In the adaptation part, the surrogate model is adapted to update task-specific parameters using both the learned initial model parameters $\theta^{ml}$ and the sampled initial dataset $\mathcal{D}_{t}$. Note that any continuously differentiable surrogate model can be integrated into the proposed framework, and any existing EA can serve as the optimizer in \texttt{MLDDEO}. In this paper, we apply well-known GPR and NN as surrogate models, along with common EAs such as covariance matrix adaptation evolution strategy (CMA-ES)~\cite{HansenO01}, particle swarm optimization (PSO)~\cite{KennedyE95}, and differential evolution (DE)~\cite{StornP97}. The pseudo codes of \texttt{MLDDEO} and \texttt{MLBO} are given in Algorithm~\ref{alg:MLDDEO} and Algorithm~\ref{alg:MLBO}, respectively. More details of the important steps are summarized below:

\begin{figure*}[t!]
    \centering
    \includegraphics[width=.8\textwidth]{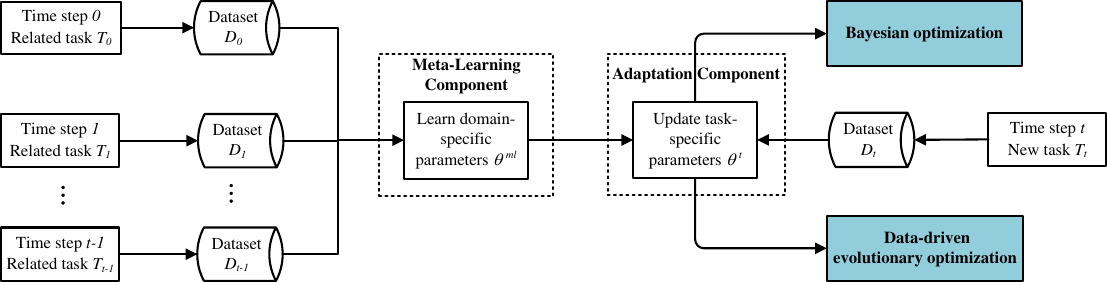}
    \caption{Flowchart of our proposed meta-learning-based optimization approach for expensive dynamic optimization.}
    \label{fig:flowchart}
\end{figure*}

\begin{itemize}
\item Initialization (line 1-4 in \texttt{MLDDEO} and line 1-3 in \texttt{MLBO}): This step involves the surrogate model parameters initialization, the training dataset initialization, and the population initialization\footnote{The population initialization is present in \texttt{MLDDEO} but absent in \texttt{MLBO}.}.
\item Meta-learning (line 11 in \texttt{MLDDEO} and line 10 in \texttt{MLBO}) and adaptation (line 15 in \texttt{MLDDEO} and line 13 in \texttt{MLBO}): The \texttt{meta-learning} component is the process of learning experience $\theta^{ml}$ and the \texttt{adaptation} component is responsible for the use of the learned experience $\theta^{ml}$.
\item Identify solution(s) (line 17 in \texttt{MLDDEO} and line 14 in \texttt{MLBO}): This step aims to pick up $\xi\ge1$ promising solutions and evaluate them by using the fitness function. These newly evaluated solutions are then used to update the training dataset $\mathcal{D}_{t}$ for the next iteration (line 18 in \texttt{MLDDEO} and line 15 in \texttt{MLBO}). In the proposed framework, the methods of selecting solutions in \texttt{MLDDEO} and \texttt{MLBO} are different. In \texttt{MLBO}, the optimization of an acquisition function across the entire search space is performed to identify the optimal solution. In \texttt{MLDDEO}, one or more solutions are selected from both the parent and offspring populations of the current generation, depending on the surrogate model employed. Specifically, when using a GPR as the surrogate model, the solution with the optimal acquisition function value
is selected. In the case of utilizing an NN as the surrogate model, a batch of top $\xi$ solutions predicted using NN is selected. 
\end{itemize}

\begin{algorithm}[t!]
\small
\caption{MLDDEO}
\label{alg:MLDDEO}
\KwIn{DOP, $f(\mathbf{x}, t)$; initial population size, $N$; initial training dataset size, $N_{I}$; promising solution(s) size, $\xi$; learning rate, $\alpha$, $\beta$}
\KwOut{Optima found at all time steps, $\mathcal{X}^{best}$}
Initialize $t\leftarrow 0$, $\mathcal{D}\leftarrow \emptyset$, $\mathcal{X}^{best}\leftarrow \emptyset$\;
Initialize surrogate model parameters $\theta^{0}$, $\theta^{ml}$\;
Initialize an evolutionary parent population $\mathcal{P}\leftarrow \{\mathbf{x}^{i}_{t}\}_{i=1}^{N}$\;
Initialize training dataset for the surrogate model $\mathcal{D}_{0}\leftarrow \{\mathbf{x}^{i}_{t}, f(\mathbf{x}^{i}_{t},t)\}_{i=1}^{N_{I}}$\;
Set the number of function evaluations $N_{t}\leftarrow N_{I}$\;
\While{termination criterion is not met}{ 
   \If{Change detected}
   {
       $\mathcal{D}\leftarrow \mathcal{D}\cup \mathcal{D}_{t}$\; $\mathcal{X}^{best}\leftarrow \mathcal{X}^{best}\cup \{\mathop{\arg\min}\limits_{\mathbf{x}_{t}\in \mathcal{D}_{t}} f(\mathbf{x}_{t},t)\}$\;
       $t\leftarrow t+1$\;
       $\theta^{ml}\leftarrow \texttt{Meta-learning}(\mathcal{D},\alpha,\beta)$\;
       Sample an initial training dataset for the new environment $\mathcal{D}_{t}\leftarrow \{\mathbf{x}^{i}_{t}, f(\mathbf{x}^{i}_{t},t)\}_{i=1}^{N_{I}}$\;
       Initialize an evolutionary parent population $\mathcal{P}\leftarrow \{\mathbf{x}^{i}_{t}\}_{i=1}^{N}$ for the new environment\;
    }
    
    $\theta^{t} \leftarrow \texttt{Adaptation}(\mathcal{D}_{t}, \theta^{ml}, \theta^{t}$)\;
    Reproduce an offspring population $\mathcal{Q}\leftarrow \{\mathbf{\widehat{x}}^{i}_{t}\}_{i=1}^{N}$\;
    Identify $\xi$ solution(s) $\bar{\mathbf{x}}\in \mathcal{P}\cup \mathcal{Q}$ and evaluate the fitness value(s) of $\bar{\mathbf{x}}$, $f(\bar{\mathbf{x}},t)$\;       $\mathcal{D}_{t}\leftarrow \mathcal{D}_{t}\cup\{\bar{\mathbf{x}},f(\bar{\mathbf{x}},t)\}$\;
    $N_{t}\leftarrow N_{t}+\xi$\;
    Conduct an environmental selection upon $\mathcal{P}\cup \mathcal{Q}$ to select the next parent population $\mathcal{P}$\; 
}
\Return $\mathcal{X}^{best}$\
\end{algorithm}

\begin{algorithm}[t!]
\small
\caption{MLBO}
\label{alg:MLBO}
\KwIn{DOP, $f(\mathbf{x}, t)$; initial training dataset size, $N_{I}$; promising solution(s) size, $\xi$; learning rate, $\alpha$, $\beta$}
\KwOut{Optima found at all time steps, $\mathcal{X}^{best}$}
Initialize $t\leftarrow 0$, $\mathcal{D}\leftarrow \emptyset$, $\mathcal{X}^{best}\leftarrow \emptyset$\;
Initialize surrogate model parameters $\theta^{0}$, $\theta^{ml}$\;
Initialize training dataset for the surrogate model $\mathcal{D}_{0}\leftarrow \{\mathbf{x}^{i}_{t}, f(\mathbf{x}^{i}_{t},t)\}_{i=1}^{N_{I}}$\;
Set the number of function evaluations $N_{t}\leftarrow N_{I}$\;
\While{termination criterion is not met}{ 
   \If{Change detected}
   {
       $\mathcal{D}\leftarrow \mathcal{D}\cup \mathcal{D}_{t}$\; $\mathcal{X}^{best}\leftarrow \mathcal{X}^{best}\cup \{\mathop{\arg\min}\limits_{\mathbf{x}_{t}\in \mathcal{D}_{t}} f(\mathbf{x}_{t},t)\}$\;
       $t\leftarrow t+1$\;
       $\theta^{ml}\leftarrow \texttt{Meta-learning}(\mathcal{D},\alpha,\beta)$\;
       Sample $N_{I}$ initial training dataset for the new environment $\mathcal{D}_{t}\leftarrow \{\mathbf{x}^{i}_{t}, f(\mathbf{x}^{i}_{t},t)\}_{i=1}^{N_{I}}$\;
    }
    
    $\theta^{t} \leftarrow \texttt{Adaptation}(\mathcal{D}_{t}, \theta^{ml}, \theta^{t}$)\;
    Identify $\xi$ solution(s) ${\mathbf{\bar{x}}}$ in the search space and evaluate the fitness value(s) of ${\mathbf{\bar{x}}}$, $f(\mathbf{\bar{x}},t)$\;       $\mathcal{D}_{t}\leftarrow \mathcal{D}_{t}\cup\{\mathbf{\bar{x}},f(\mathbf{\bar{x}},t)\}$\;
    $N_{t}\leftarrow N_{t}+\xi$\;
}
\Return $\mathcal{X}^{best}$\
\end{algorithm}

\begin{remark}
Our proposed framework offers a high degree of flexibility. Firstly, it can be combined with both evolutionary optimization and BO approaches. Secondly, it can be seamlessly integrated with any continuously differentiable surrogate model, allowing for compatibility with a wide range of models. Moreover, any existing EA can be employed as the optimizer within the framework of \texttt{MLDDEO}.
\end{remark}

\subsection{Gradient-based Meta-learning for Surrogate Modeling}
\label{sec:meta-surrogate}
Parameter optimization is vital in the surrogate modeling~\cite{YangS20}. It is widely recognized that the configuration of initial parameters significantly influences the outcome of parameter optimization. In the conventional data-driven evolutionary optimization and BO approaches, the initial parameters are typically determined using rules of thumb or randomly generated by an experimental design method. However, considering the potential relevant patterns of a DOP across different environments, valuable information can be collected from previously optimized problems~\cite{YazdaniCYBJY21A,HuZJYZ23}. The effective utilization of this knowledge can facilitate the process of surrogate modeling for a new environment. In addition, this becomes crucial when considering an expensive optimization scenario, where the computational budget is severely restricted at each time step. To address the above issue, this section elaborates on a gradient-based meta-learning approach~\cite{FinnAL17} for surrogate modeling to learn experience from previously optimized time steps of an expensive DOP.

\begin{algorithm}[t!]
\small
\caption{Gradient-based Meta-learning for Surrogate Modeling}
\label{alg:GML}
\KwIn{Meta-training dataset, $\mathcal{D}$; learning rate, $\alpha$, $\beta$}
\KwOut{Model parameters $\theta^{ml}$}
Initialize model parameters $\theta$\;
\While{termination criterion is not met}
   {
    Sample batch of time steps $\{T_i\}_{i=1}^{m}$ from all previously optimized time steps $\{T_r\}_{r=0}^{t-1}$\;
    \For{all $T_i$}
      {
       Sample support set $\mathcal{D}_{i}^{s}$ from the dataset of $T_i$\;
       Compute loss function $\mathcal{L}(\theta, \mathcal{D}_{i}^{s})$ according to the surrogate model used\;
       Compute the $\theta'_{i}$ with gradient descent in~\pref{eq:inner}\;
       Sample query set $\mathcal{D}_{i}^{q}$ from the dataset of $T_{i}$\;
       Compute loss function $\mathcal{L}(\theta'_{i}, \mathcal{D}_{i}^{q})$ according to the surrogate model used\;
      }
     Compute meta-loss function $\mathcal{L}_{meta}(\theta'_{b},\mathcal{D}_{b}^{q})$ in \pref{eq:loss}\;
     Update $\theta$ using meta-optimizer with respect to $\theta$ in \pref{eq:outer}\;
   }
   $\theta^{ml}\leftarrow \theta$\;
\Return $\theta^{ml}$\
\end{algorithm}

Whenever an environment changes, we employ the \texttt{meta-learning} component to leverage historical data to learn good initial model parameters for the new environment. This component encompasses both inner optimization and outer optimization. In a nutshell, the inner optimization aims to quickly adjust the model parameters to fit the requirements of the problem at each time step, while the outer optimization focuses on learning shared patterns among the problems at different time steps to obtain domain-specific parameters. The working strategy of the \texttt{meta-learning} component is given in Algorithm~\ref{alg:GML}. First, $m$ time steps are randomly sampled from all previously optimized time steps $\{T_{r}\}_{r=0}^{t-1}$ (line 3), each of which contains $K$ observations randomly drawn from the corresponding dataset to constitute the support set $\mathcal{D}_{i}^{s}=\{\mathbf{x}_{i}^{k},f(\mathbf{x}_{i}^{k},i)\}_{k=1}^{K}$ (line 5) and query set $\mathcal{D}_{i}^{q}=\{\mathbf{x}_{i}^{k},f(\mathbf{x}_{i}^{k},i)\}_{k=1}^{K}$ (line 8), $i\in{{1,\cdots,m}}$, $k\in{{1,\cdots, K}}$, respectively. In the inner optimization, the parameters $\theta$ of the surrogate model are computed using one gradient update (line 7):
\begin{equation}
    \begin{aligned}
        \theta'_{i}=\theta-\alpha\nabla_{\theta}\mathcal{L}(\theta,\mathcal{D}_{i}^{s}) \\
    \end{aligned},
    \label{eq:inner}
\end{equation}
where $\alpha$ is the inner-loop learning rate and $\nabla_{\theta}\mathcal{L}(\theta,\mathcal{D}_{i}^{s})$ is the gradient of the loss function in regard to the support set $\mathcal{D}_{i}^{s}$.

In the outer optimization, the model parameters can be updated by optimizing the meta-loss function of $\theta'_{i}$ with respect to $\theta$ across the query sets $\mathcal{D}_{i}^{q}$ defined as (line 11):
\begin{equation}
    \begin{aligned}
        \mathcal{L}_{meta}(\theta'_{b},\mathcal{D}_{b}^{q})=\sum_{i=1}^{m}\mathcal{L}(\theta'_{i},\mathcal{D}_{i}^{q})\\
    \end{aligned},
    \label{eq:loss}
\end{equation}
where $\mathcal{L}_{meta}(\theta'_{b},\mathcal{D}_{b}^{q})$ is the loss of the $b$-th batch.

Accordingly, the parameters $\theta$ are updated in the outer optimization using the meta-optimizer (line 12). For example, if the gradient descent is used as the meta-optimizer, the model parameters $\theta$ are updated, which is given by:
\begin{equation}
    \begin{aligned}
        \theta=\theta-\beta\nabla_{\theta}\mathcal{L}_{meta}(\theta'_{b},\mathcal{D}_{b}^{q})\\
    \end{aligned},
    \label{eq:outer}
\end{equation}
where $\beta$ is the outer-loop learning rate. The termination criteria in the \texttt{meta-learning} component can be reaching a fixed number of epochs (line 2). Finally, $\theta$ is assigned to $\theta^{ml}$ as the output (line 14), which serves as the initial model parameters of the surrogate model for the new environment. Note that when different surrogate models are applied in the \texttt{meta-learning} component, we directly employ the corresponding loss function to replace the loss function in line 6 and line 9 of Algorithm~\ref{alg:GML}. For example, the loss function of GPR is considered as maximizing the log marginal likelihood~\cite{RasmussenW06}. For NN, the minimum mean-squared error~\cite{HastieTH09} is used as the loss function. 

\begin{remark}
The \texttt{meta-learning} component is capable of learning the experience from the previously optimized time steps of an expensive DOP when a change occurs. Since the meta-training dataset used is collected from these optimized time steps, the training process of the \texttt{meta-learning} component does not incur extra evaluation costs. Furthermore, this component enables the optimization process to be efficiently initiated in the new environment, facilitating adaptation to the dynamic changing environment.
\end{remark}

\subsection{Adaptation for new environment}
\label{sec:Adaptation}
In general, the experience learned by the \texttt{meta-learning} component, denoted as $\theta^{ml}$, represents the domain-specific features of the related tasks. To optimize a new task, the surrogate model needs to update the task-specific parameters with the guidance of $\theta^{ml}$. Bearing this consideration in mind, we develop a \texttt{adaptation} component to carry out this process. 

\begin{algorithm}[t!]
\small
\caption{Adaptation}
\label{alg:Adaptation}
\KwIn{A dataset sampled from new task $T_{t}$, $\mathcal{D}_{t}$; surrogate model parameters obtained from meta-learning, $\theta^{ml}$; current surrogate model parameters, $\theta^{t}$}
\KwOut{Updated surrogate model parameters, $\theta^{t}$}
\If{Change detected}
{
    $\theta^{t}\leftarrow \theta^{ml}$\; 
}
\While{termination criterion is not met}
{ 
    Compute the loss function according to the surrogate model used based on $\theta^{t}$ and $\mathcal{D}_{t}$\;
    Update $\theta^{t}$ using the optimizer\;
}
\Return $\theta^{t}$\
\end{algorithm}

The algorithmic implementation of the \texttt{adaptation} component is shown in Algorithm~\ref{alg:Adaptation}. It contains the following steps. Firstly, the inputs of the \texttt{adaptation} component include the dataset $\mathcal{D}_{t}$ sampled from the new task $T_t$, the experience $\theta^{ml}$ obtained from the \texttt{meta-learning} component, and the current surrogate model parameters $\theta^{t}$. If a change is detected, the surrogate model parameters $\theta^{t}$ of the new task $T_t$ are initialized with experience $\theta^{ml}$ (line 1-3); otherwise, the current model parameters will be retained. Last but not least, the surrogate model is updated using the parameters $\theta^{t}$ and the dataset $\mathcal{D}_{t}$ (lines 5-6). For the sake of algorithmic rationality, the optimizer employed for updating the surrogate model in the \texttt{adaptation} component is consistent with the meta-optimizer utilized in the \texttt{meta-learning} component. The termination criteria is to reach the convergence condition of the adopted optimizer, such as reaching the maximum number of iterations or achieving the predetermined error requirement (line 4).

\subsection{Analysis of Computational Complexity}
\label{sec:Computational_Complexity}
The computational complexity of the \texttt{meta-learning} component is determined by the number of tasks, epochs, and the complexity of the surrogate model. For GPR, the computational complexity of the \texttt{meta-learning} component can be expressed as $\mathcal{O}(I_{ml}mK^3)$. Here, $I_{ml}$ is the number of epochs of \texttt{meta-learning} component, $m$ is the number of tasks per epoch, and $K$ is the size of the support set and the query set. For NN, the computational complexity of the \texttt{meta-learning} component can be expressed as $\mathcal{O}(I_{ml}LN_{e}^2mK)$. Here, $L$ represents the number of layers, and $N_{e}$ represents the number of neurons per layer.

The computational complexity of data-driven evolutionary optimization depends on the surrogate model and the corresponding baseline EA. The training complexities of the GPR and NN models are $\mathcal{O}(N_{t}^3)$ and $\mathcal{O}(LN_{e}^2N_{t})$, respectively. Here, $N_{t}$ represents the number of training points in the surrogate model. The evaluation complexity of SACMA-ES with UCB is $\mathcal{O}(N_{t}n)$, where $n$ is the number of dimensions in the search space. The evaluation complexity of SAPSO and SADE with UCB is $\mathcal{O}(NN_{t})$, where $N$ is the population size. Therefore, the overall computational complexity of SACMA-ES, SAPSO, and SADE is primarily determined by the model training process. Using GPR, the computational complexity is $\mathcal{O}(I_{eo}N_{t}^3)$, while using NN, it is $\mathcal{O}(I_{eo}LN_{e}^2N_{t})$, where $I_{eo}$ represents the number of iterations of the data-driven evolutionary optimization approach.

The computational complexity of BO is determined by both GPR training and the acquisition function optimization. Typically, the computational complexity of training of the GPR model is still $\mathcal{O}(N_{t}^3)$. For the UCB used in this study, the computational complexity of optimizing the acquisition function is $\mathcal{O}(nN_{t}^2)$ with $N_{t}$ data points and $n$ dimensions. The computational complexity of BO is typically $\mathcal{O}(I_{BO}N_{t}^3)$, where $I_{BO}$ is the number of iterations of BO.

In summary, the overall computational complexity of the proposed algorithm framework can be categorized into the following three cases: 1) The complexity of \texttt{MLBO} is $\mathcal{O}(I_{ml}mK^3)$ + $\mathcal{O}(I_{BO}N_{t}^3)$. 2) The complexity of \texttt{MLDDEO} with GPR as the surrogate model is $\mathcal{O}(I_{ml}mK^3)$ + $\mathcal{O}(I_{eo}N_{t}^3)$. 3) The complexity of \texttt{MLDDEO} with NN as the surrogate model is $\mathcal{O}(I_{ml}LN_{e}^2mK)$ + $\mathcal{O}(I_{eo}LN_{e}^2N_{t})$. In terms of complexity, the computational complexity of the proposed framework is significantly influenced by the surrogate model, and the GPR model entails a higher complexity compared to the NN considered in this paper.

\section{Experimental Setup}
\label{sec:experiment_Setup}

The experimental setup is presented in this section, including benchmark test problems, peer algorithms, performance metrics, statistical test methods, and parameter settings.

\subsection{Benchmark Problems and Peer Algorithms}
According to~\cite{YazdaniOCBNY22}, we consider the moving peaks benchmark (MPB)~\cite{Jurgen99} in this empirical study, the most widely used synthetic problem in the DOP field. The problem in this benchmark consists of a landscape with a definable quantity of peaks. When the environment changes, the height, width, and location of each peak are varied. MPB can generate scalable objective functions with a configurable number of peaks. Each peak is capable of being the global optimum at the current environment. The mathematical definition of MPB is described in Appendix A of the supplemental material\footnote{The supplementary material can be downloaded from the following address: \href{https://github.com/HuanZhang0320/Document/raw/main/Supplementary\%20Material.pdf}{Supplementary material}} due to the page limitation.

To validate the performance of the proposed algorithms, six data-driven optimization algorithms and three evolutionary optimization algorithms are considered for comparison, including \texttt{DIN}~\cite{MoralesB15}, \texttt{MBO}~\cite{RichterSCRL20}, \texttt{TBO}~\cite{ChenL21}, \texttt{DETO}~\cite{LiCY23}, restart BO \texttt{(RBO)}, vanilla BO \texttt{(CBO)}, and restart data-driven evolutionary optimization \texttt{(RDDEO)}. Among them, \texttt{RBO}, \texttt{CBO}, and \texttt{RDDEO} are variants of the proposed algorithm in this paper. In particular, \texttt{RBO} and \texttt{RDDEO} are dynamic versions of BO and DDEO with a random reinitialization strategy, that is, restart the optimization process from scratch if the change occurs. This paper applies three iconic EAs, i.e., CMA-ES, PSO, and DE. Accordingly, there are three SAEAs, i.e., \texttt{SACMA-ES}, \texttt{SAPSO}, and \texttt{SADE}. In \texttt{CBO}, it neglects the dynamic characteristics of DOP and instead employs all data collected so far to train the surrogate model. The working mechanisms of other peer algorithms can be found in Appendix A of the supplemental material.

\subsection{Performance Metrics and Statistical Tests}
In the experimental studies, we adopt two performance metrics to evaluate the performance of different algorithms.

\emph{1) Best error before change ($E_{BBC}$)}~\cite{YazdaniCYBJY21}: it is commonly used performance metric in the field~\cite{YazdaniCYBJY21}. $E_{BBC}$ calculates the last error at the end of each environment. 
\begin{equation}
\begin{aligned}
    E_{BBC}=\frac{1}{T}\sum_{t=1}^{T}[f(\mathbf{x}^{\star},t)-f(\mathbf{x}^{best},t)]
\end{aligned},
\label{eq:EBBC}
\end{equation}
where $T$ is the number of environments, $(\mathbf{x}^{\star},t)$ is the global optimum at the $t$-th environment, and $(\mathbf{x}^{best},t)$ is the best solution found at the end of the $t$-th environment.

\emph{2) Budget ratio ($\rho_c$)}~\cite{LiCY23}: it assesses the ratio of the computational budget required between the best algorithm and a peer algorithm at each time step. 
\begin{equation}
\begin{aligned}
    \rho_c=\frac{1}{T}\sum_{t=1}^{T}\frac{N^{ts}_{FE}}{\widetilde{N}^{t*}_{FE}}
\end{aligned},
\label{eq:rho}
\end{equation}
where $\widetilde{N}^{t*}_{FE}$ represents the number of FEs required by the best algorithm at the $t$-th time step to achieve its best solution found $f(\widetilde{x}^{\ast}, t)$. $N^{ts}_{FE}$ is the number of FEs consumed by one of the other comparative algorithms to obtain the same solution. In practice, $\rho_c \ge 1$ and the larger $\rho_c$ becomes, the poorer the performance of the corresponding peer algorithm.

To obtain a statistical interpretation of the significance between comparison results, three statistical measures are applied in our empirical study, including Wilcoxon rank-sum test~\cite{Wilcoxon45}, Scott-Knott test~\cite{MittasA13}, and $A_{12}$ effect size~\cite{VarghaD00}. The basic idea of three statistical measures is provided in Appendix A of the supplemental material due to space limitations.

\subsection{Parameter Settings}
\label{sec:PSs}
In our empirical study, the parameters for the compared algorithms are kept the same as in the published papers. Some key parameters are listed below.

\emph{1) Settings of test problems}: The number of decision variables is set as $n\in\{4, 6, 8, 10\}$. The parameters for the MPB problem are given in Table I of the supplementary material. 

\emph{2) Settings of computational budget}: In our experiment, the total number of FEs is fixed, and the number of FEs per environment ($N_{FE}^{t}$) is also limited. For fairness, the initial sampling size is set to $4\times n$ and the maximum number of FEs is set to $5\times n$ for all algorithms. When calculating $\rho_c$, the number of FEs per environment can reach up to $7\times N_{FE}^{t}$, regardless of whether the corresponding comparative algorithm acquires $f(\tilde{x}^{\ast}, t)$. Once the maximum number of FEs per environment is reached, the environment changes. The experiment terminates when the total computational cost is exhausted.

\emph{3) Settings of surrogate models}: We use the Python packages GPy~\cite{gpy12} and scikit-learn~\cite{PedregosaVGMTGBPWDVPCBPD11} to implement GPR and NN, respectively. For GPR, the initial model parameters are set to their default values in GPy: $l=1$, $\sigma^{2}_{f}=1$, and $\sigma^{2}_{n}=0.01$. For NN, we implement a model with three hidden layers of size 40 and ReLU nonlinearities using the suggested parameters in scikit-learn. Additionally, a batch size of $\xi=5$ is used when selecting promising solutions for NN.

\emph{4) Settings of meta-learning}: The number of few-shot samples is set as $K=5$. The number of epochs is set as $I_{ml}=20000$. The inner-loop learning rate is set as $\alpha=0.01$. The learning rate in the meta-optimizer (outer-loop learning rate) is also set as $\beta=0.01$. For fairness the initial parameters of the meta-learning are consistent with the other algorithms.

\emph{5) Settings of acquisition function}: The upper confidence bound (UCB)~\cite{SrinivasKKS10} is employed as the acquisition function in this paper. It is implemented by the Python packages GPyOpt~\cite{gpyopt16}. The parameter $w$ for UCB is set to 2, consistent with the default setting in GPyOpt. 

\emph{6) Number of environments}: There are 10 changes (environments) for each run in our experiment.

\emph{7) Number of repeated runs}: Each algorithm performs 20 times independently with different random seeds.

\section{Empirical Results}
\label{sec:empirical_results}
In this section, we provide a comprehensive study of the proposed framework by investigating the following concerns \footnote{Due to the page limitation, the study in the manuscript is focused on three concerns, while the remaining investigations are provided in Appendix B of the supplementary material. The additional investigations encompass discussing the proposed algorithm's performance under dynamic changes in peak height and center, conducting a sensitivity study of the algorithm parameters, and exploring the performance of the proposed framework when employing different optimization solvers and surrogate models.}: 1) the effectiveness of the solution quality of the proposed algorithm, 2) the efficiency of the computational cost of the proposed algorithm, 3) how does the proposed algorithm work?

\subsection{Effectiveness of the proposed algorithm}
\label{sec:effectiveness}
In this subsection, we investigate the performance of the proposed framework \texttt{MLO} against the state-of-the-art algorithms. To answer concern 1, the experimental results are analyzed in four parts, i.e., Wilcoxon rank sum test, loss function curves, Scott-Knott test, and $A_{12}$ effect size. 

Based on the Wilcoxon rank sum test, the statistical comparison results of $E_{BBC}$ values are shown in Table II and III in Appendix B of the supplementary material. From these results, it is evident that the proposed algorithm instances perform significantly better in $E_{BBC}$ metric than the other peer algorithms in all comparisons.

\begin{figure*}[t!]
    \centering
    \includegraphics[width=.46\textwidth]{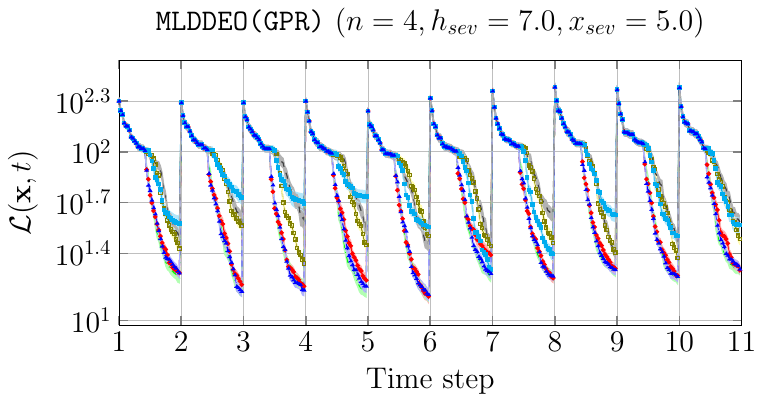}
    \includegraphics[width=.46\textwidth]{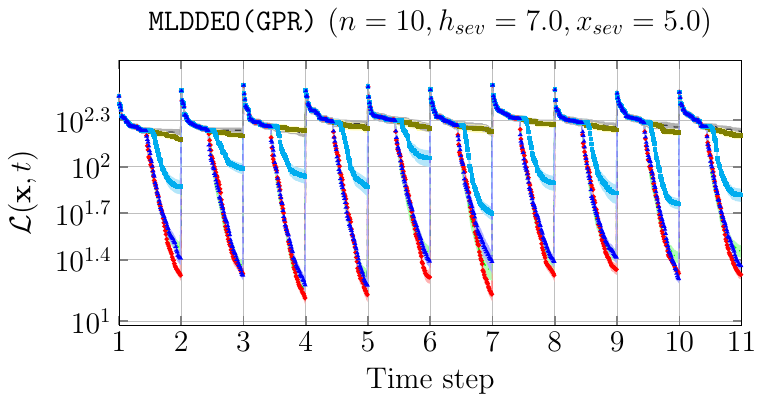}
    \vfill
    \includegraphics[width=.7\textwidth]
    {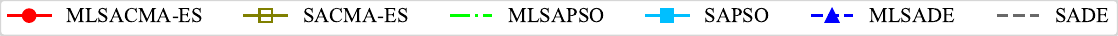}
    \vfill
    \includegraphics[width=.46\textwidth]{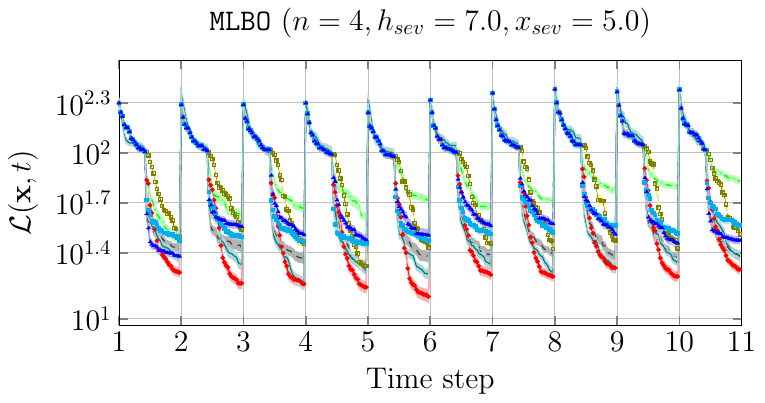}
    \includegraphics[width=.46\textwidth]{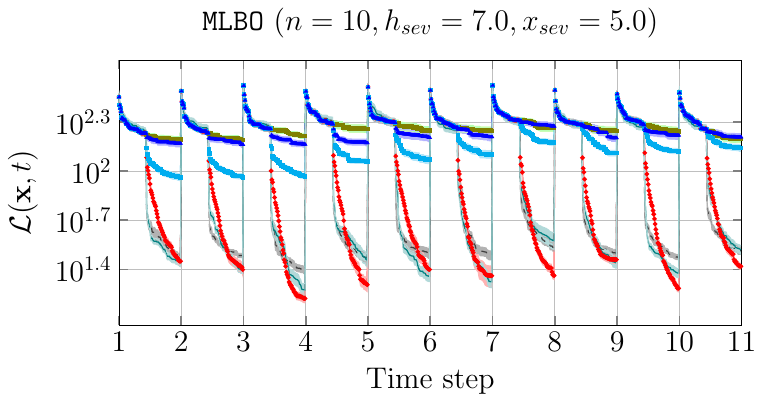}
    \includegraphics[width=.6\textwidth]
    {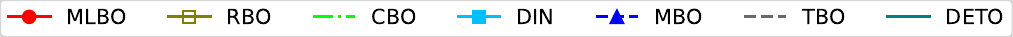}
    \vfill
    \caption{Loss function of the mean errors between the true objective functions and the best objective values along with a confidence level over time at different dimensions with $h_{sev}=7.0$ and $x_{sev}=5.0$ when comparing \texttt{MLDDEO} by using GPR and \texttt{MLBO} with other peer algorithms, respectively.}
    \label{fig:MLO-GPR-mpb-loss-s=5h=7}
\end{figure*}

To intuitively assess the solutions obtained by each algorithm during the evolutionary process, we record the loss function defined as follows:
\begin{equation}
    \begin{aligned}
        \mathcal{L}(\mathbf{x},t)=f(\mathbf{x}^{\star},t)-f(\mathbf{\widetilde{x}},t)
        \end{aligned},
\end{equation}
where $(\mathbf{\widetilde{x}},t)$ represents the best solution discovered at the corresponding FE of the $t$-th environment, $(\mathbf{x}^{\star},t)$ is the global optimum at the $t$-th environment. The loss function curves obtained by different algorithms over time are shown in Fig.~\ref{fig:MLO-GPR-mpb-loss-s=5h=7}. The complete results are available in Figs. 17 in Appendix B of the supplemental  material. From these plots, it evidently shows that \texttt{MLO} obtains solutions with smaller errors from the global optimum than peer algorithms in most time steps at different dimensions. In addition, we can find that the superiority of \texttt{MLO} becomes more pronounced compared to peer algorithms with the increase of dimensionality. In particular, \texttt{SACMA-ES}, \texttt{SAPSO}, \texttt{SADE}, and \texttt{RBO} all restart from scratch, while \texttt{CBO} simply ignores the environmental changes. This makes it difficult for them to adapt to the dynamic environment with a strictly restricted computational budget. \texttt{MBO} claimed to include the time as a covariate and learn the influence of the time. Nevertheless, its performance crashes to almost the same level as \texttt{RBO} when $n=10$. Both \texttt{DIN} and \texttt{TBO} introduce an improved covariance function to leverage the previous information. However, they assume earlier collected information is less informative for the current environment, resulting in the possibility of overlooking useful information. \texttt{DETO}, derived from \texttt{TBO}, exhibits a certain level of competitiveness. In contrast, the proposed framework can leverage meta-learning to learn a priori, whether for BO or evolutionary optimization. This enables a rapid initiation of the optimization process in the new environment. 

\begin{figure*}[t]
    \centering
    \includegraphics[width=.43\textwidth]{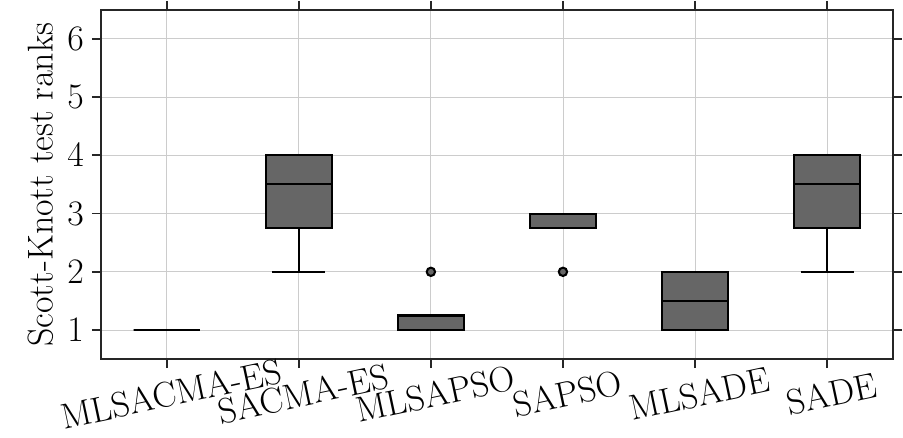}
    \includegraphics[width=.43\textwidth]{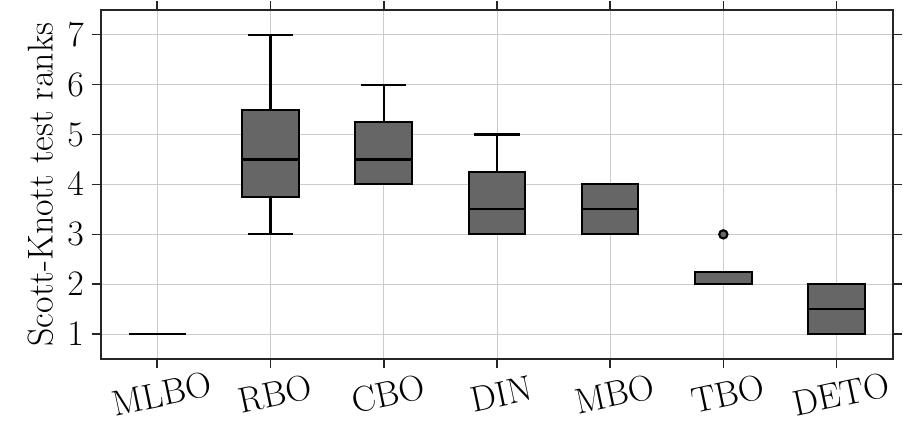}
    \caption{Box plots of Scott-Knott test ranks of $E_{BBC}$ obtained by the proposed algorithm instances and the corresponding peer algorithms with $h_{sev}=7.0$ and $x_{sev}=5.0$ (the smaller rank is, the better performance achieved).}
    \label{fig:sk-concern1}
\end{figure*}

To facilitate a clear analysis of ranking among different algorithms, the Scott-Knott test is employed to partition their metric values into different groups. It will be chaotic to list all ranking results on different dimensions among these algorithms. Therefore, all the Scott-Knott test results are pulled together and the distribution and median are presened as box plots in Fig.~\ref{fig:sk-concern1}. As shown in Fig.~\ref{fig:sk-concern1}, the proposed framework achieves the smallest summation rank across all comparisons both for \texttt{MLDDEO} and \texttt{MLBO}. It is further confirmed that the proposed algorithm is the best algorithm in the corresponding comparisons. 

\begin{figure*}[t!]
    \centering
    \includegraphics[width=.46\textwidth]{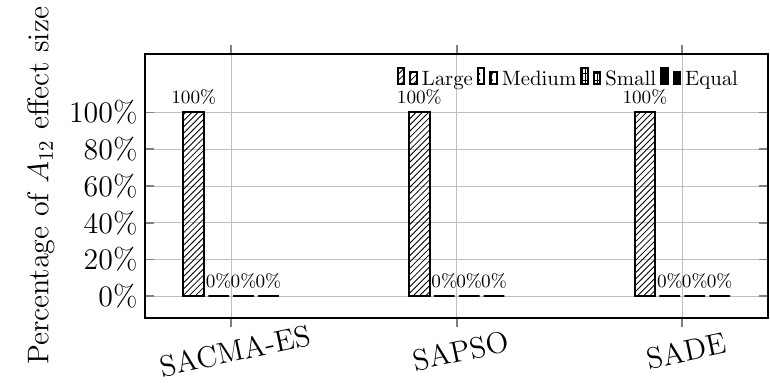}
    \includegraphics[width=.46\textwidth]{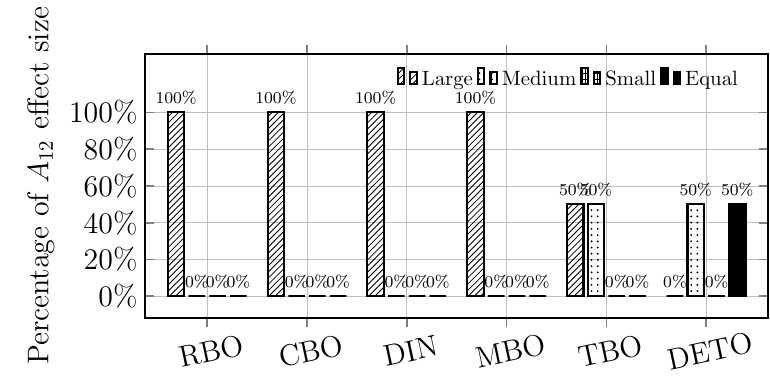}
    \caption{Percentage of $A_{12}$ effect size of $E_{BBC}$ with $h_{sev}=7.0$ and $x_{sev}=5.0$ when comparing \texttt{MLDDEO} or \texttt{MLBO} with corresponding state-of-the-art peer algorithms.}
    \label{fig:a12-concern1}
\end{figure*}

In addition, we evaluate the $A_{12}$ effect size to compare the proposed algorithm instances with the corresponding peer algorithms under two optimization mechanisms separately. $A_{12}$ is a pairwise performance comparison, all the results are pulled together again and the percentage of the large, medium, small, and equivalent effects size is calculated. The statistical results given in Fig.~\ref{fig:a12-concern1} further validate the remarkable advantage of the proposed framework. Specifically, the percentage of the large effect sizes is all $100\%$ except for the comparison with \texttt{TBO} and \texttt{DETO}. Nevertheless, \texttt{MLBO} outperforms \texttt{TBO} by $50\%$ in both large and medium effect size. Additionally, \texttt{MLBO} outperforms \texttt{DETO} by $50\%$ in medium effect size, with $50\%$ of the comparisons being statistically equivalent.

Due to the page limitation, the remaining experimental results of using NN as the surrogate model in response to concern 1 can be found in Appendix B of the supplementary material, further supporting the effectiveness of the proposed algorithm instances.

\begin{framed}
    \underline{Response to concern 1:} We have the following takeaways from the experiments: In terms of effectiveness, all algorithm instances within the proposed framework have consistently exhibit significantly superior performance against the peer algorithms when given a limited number of evaluations. This demonstrates the effectiveness of the solution quality achieved by the proposed algorithm. 
\end{framed}

\subsection{Efficiency of the proposed algorithm}
\label{sec:efficiency}
The training process of the \texttt{meta-learning} component, as stated in Remark 2, does not consume FEs. However, it does require extra runtime. To address concern 2, we discuss the computational cost of all algorithms in terms of both evaluation cost and time cost in this subsection.

\begin{figure*}[t!]
    \centering
    \includegraphics[width=.46\textwidth]{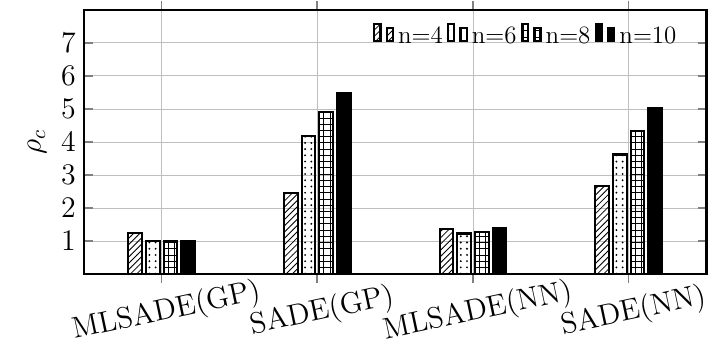}
    \includegraphics[width=.46\textwidth]{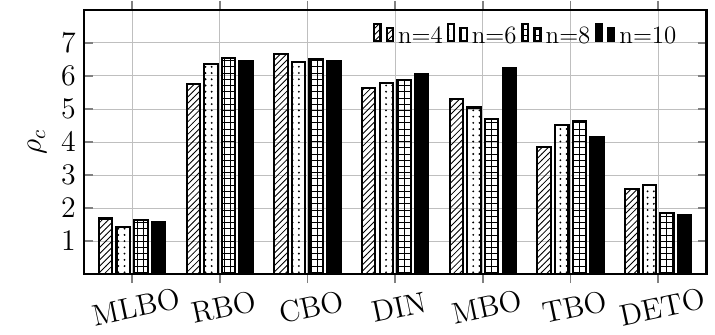}
    \caption{Bar chart of $\rho_c$ achieved by \texttt{MLO} and the other comparative algorithms.}
    \label{fig:pc-concern2}
\end{figure*}

To assess the evaluation cost among different algorithms, we compare the computational budget consumed when each algorithm achieves the same solution. We choose \texttt{MLBO}, \texttt{MLSADE(GP)}, and \texttt{MLSADE(NN)} as representative algorithms from each optimization mechanism and surrogate model. The bar chart of $\rho_c$ is shown in Fig.~\ref{fig:pc-concern2}. From these plots, we can see that the proposed algorithms obtain smaller values of $\rho_c$ under both optimization mechanisms. This indicates that the proposed algorithms require fewer evaluations compared to the peer algorithms to achieve the same solution. In expensive optimization scenarios, this significantly conserves computational resources.

\begin{figure}[t!]
    \centering
    \includegraphics[width=.46\textwidth]{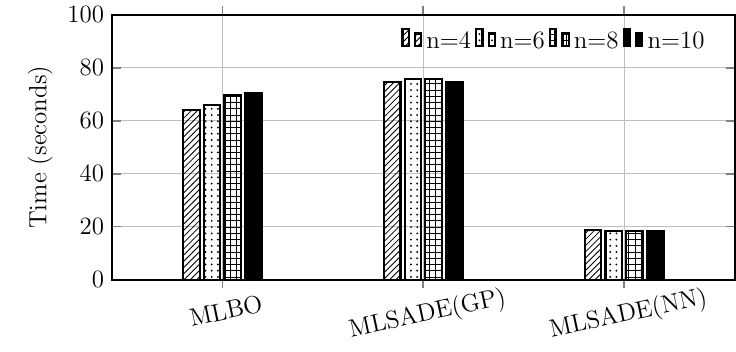}
    \caption{Collected comparisons of CPU wall clock time of \texttt{meta-learning} component on a computer with an Intel Core i9, 3.00-GHz CPU when using different the proposed algorithm instances.
    \label{fig:mltime-concern2}}
\end{figure}

\begin{figure}[t!]
    \centering
     \includegraphics[width=.46\textwidth]{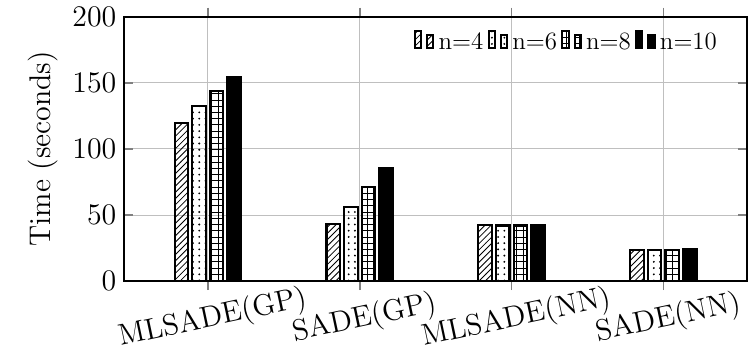}
    \includegraphics[width=.46\textwidth]{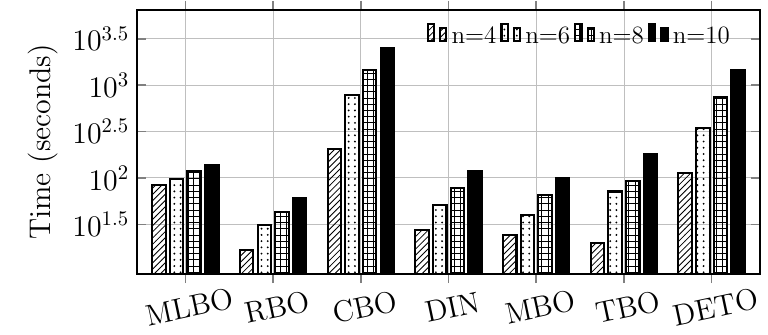}
    \caption{Collected comparisons of CPU wall clock time on a computer with an Intel Core i9, 3.00-GHz CPU when using same evaluation costs.}
    \label{fig:same-FEs-time-concern2}
\end{figure}

\begin{figure}[t!]
    \centering
    \includegraphics[width=.46\textwidth]{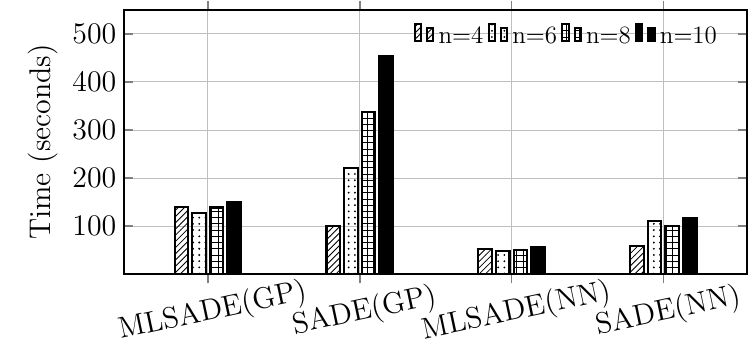}
    \includegraphics[width=.46\textwidth]{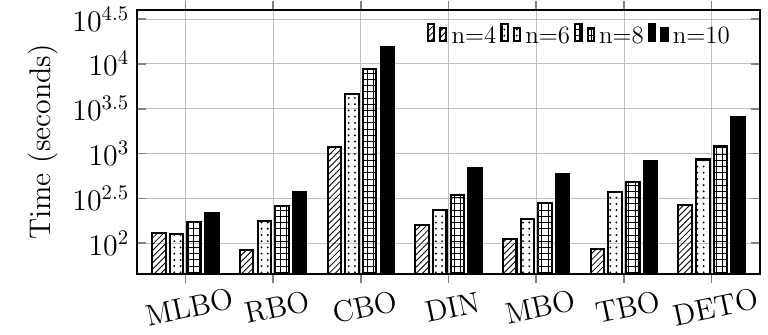}
   \caption{Collected comparisons of CPU wall clock time on a computer with an Intel Core i9, 3.00-GHz CPU when achieving the same solution.}
    \label{fig:same-solution-time-concern2}
\end{figure}

To investigate the time cost of different algorithms, we first presented the runtime of meta-learning training process across different dimensions. Furthermore, we displayed the runtime of all algorithms with same evaluation costs and the time taken by each algorithm to achieve the same solution. Fig.~\ref{fig:mltime-concern2} shows the average CPU time for a single training cycle of the \texttt{meta-learning} component. It reveals that the runtime for a training cycle of the \texttt{meta-learning} component is approximately 70 seconds for the GPR model across different dimensions considered in this paper, while for the NN model, it is just under 20 seconds. The CPU time of all algorithms with same evaluation costs is given in Fig.~\ref{fig:same-FEs-time-concern2}. From this figure, it can be observed that \texttt{RBO} incurs the least time as it restarts the optimization process from scratch, whereas \texttt{CBO} consumes the most time since it utilizes all historical data to train the surrogate model. In \texttt{TBO}, employing MOGP is more time-consuming compared to using single-output GP in \texttt{DIN} and \texttt{MBO}. Although \texttt{DETO} use hierarchical MOGP with fewer hyperparameters to enhance optimization cost-efficiency compared to conventional MOGP, the use of a hybrid EA to optimize acquisition functions incurs a significant time cost. The runtime of \texttt{MLBO} ranks in the middle among all algorithms when spending the same evaluation costs. However, as shown in Fig.~\ref{fig:same-solution-time-concern2}, it tasks the least time when achieving the same solution. Similarly, as depicted in Fig.~\ref{fig:same-FEs-time-concern2}, due to the introduction of the \texttt{meta-learning} component, the runtime required for \texttt{MLSADE(GP)} and \texttt{MLSADE(NN)} when using the same evaluation costs is higher than that for \texttt{SADE(GP)} and \texttt{SADE(NN)}. However, the situation is exactly the opposite when obtaining the same solution, as illustrated in Fig.~\ref{fig:same-solution-time-concern2}. Furthermore, it can be observed from Fig.~\ref{fig:same-solution-time-concern2} that the ranking of other peer algorithms remains unchanged compared to Fig.~\ref{fig:same-FEs-time-concern2}.

\begin{framed}
    \underline{Response to concern 2:} We derived the following insights from the empirical study: 1) The proposed framework can save computational costs in terms of both evaluation cost and time cost, thereby enhancing algorithm efficiency. 2) Regarding evaluation cost, the training process of the \texttt{meta-learning} component does not consume FEs. Furthermore, all algorithm instances within the proposed framework consume fewer FEs to obtain the same solution compared to the peer algorithms. 3) In terms of time cost, the training process of the \texttt{meta-learning} component requires additional time, but each training cycle does not exceed 70 seconds on the functions considered in this paper. Moreover, the proposed algorithm instances incur the least time cost among all corresponding peer algorithms when obtaining the same solution.
\end{framed}

\subsection{How does the proposed algorithm work?}
\label{sec: reason for work}
The empirical study conducted in the preceding two subsections has indicated the proposed framework outperforms the state-of-the-art peer algorithms. In this subsection, we aim to discover the driving force that renders the proposed algorithm instances effective. We choose \texttt{MLBO} as the baseline and analyze quantitatively some experimental examples. 

\begin{figure}[t!]
    \centering
    \subfigure[Locate at the same peak]{
    \includegraphics[width=.46\textwidth]{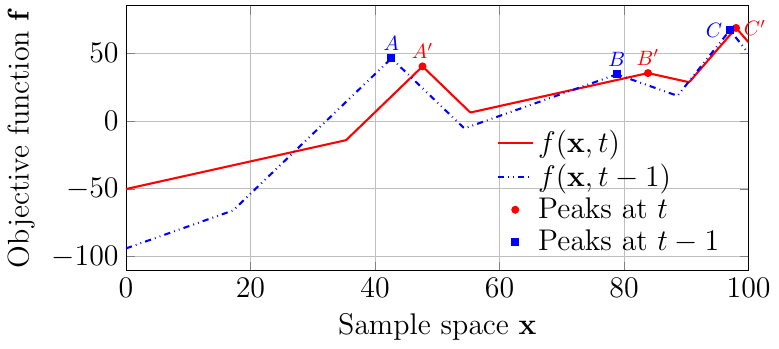}}
    \subfigure[Jump from one peak to another]{
    \includegraphics[width=.46\textwidth]{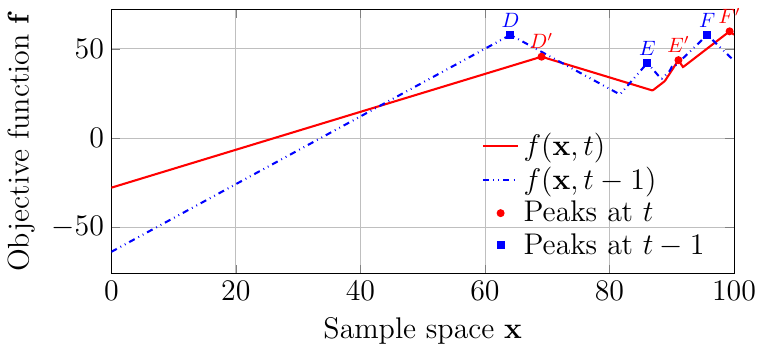}}
    \caption{Illustrative example of the global optimum of the true objective functions at adjacent environment under two scenarios.}
    \label{fig:mpb-concern3-truef}
\end{figure}

We apply MPB with $n = 1$ for better visual interpretation. Based on the dynamic characteristic of the MPB problem, we consider the following two scenarios: the global optimum of the true objective functions at adjacent environment 1) locates at the same peak. As shown in Fig.~\ref{fig:mpb-concern3-truef} (a), the three peaks at $(t-1)$-th time step are at points $A$, $B$, and $C$, corresponding to peaks $A'$, $B'$, and $C'$ at $t$-th time step, respectively. When the environment changes, the global optimum moves from the peak $C$ to the peak $C'$. 2) jumps from one peak to another. As depicted in Fig.~\ref{fig:mpb-concern3-truef} (b), the peaks $D$, $E$, and $F$ at $(t-1)$-th time step correspond to the peaks $D'$, $E'$, and $F'$ at $t$-th time step, respectively. The global optimum jumps from the peak $D$ to the peak $F'$ after the change occurs. Obviously, the second scenario is more challenging. To address concern 3, we plan to investigate the superiority of \texttt{MLBO} from two aspects, i.e., the effectiveness of the \texttt{meta-learning} component and \texttt{adaptation} component.   

\emph{1) Effectiveness of the meta-learning component:} We track two measures, average loss over sampled batches ($AL_{B}$) and model parameters values for each batch ($MP_{B}$), to study the usefulness of the \texttt{meta-learning} component. In the GPR model, $AL_{B}$ that measures the average negative log marginal likelihood over sampled batches in regard to the query set is formulated as
\begin{equation}
    \begin{aligned}
        AL_{B}=\frac{1}{B}\sum_{b=1}^{B}\mathcal{L}_{meta}(\theta'_{b},D_{b}^{q})=-\frac{1}{B}\sum_{b=1}^{B}\sum_{i=1}^{m}\log p(\mathbf{f}_{i}|\mathbf{X}_{i},\theta'_{i})
    \end{aligned},
    \label{eq:ALB}
\end{equation}
where $B$ denotes the number of sampled batches, $\mathcal{L}_{meta}(\theta'_{b},D_{b}^{q})$ is the loss of the $b$-th batch in~\pref{eq:loss} as introduced in~\pref{sec:meta-surrogate}, $\log p(\mathbf{f}_{i}|\mathbf{X}_{i},\theta'_{i})$ is the log marginal likelihood, $\mathbf{f}_{i}=\{f_{i}^{k}\}_{k=1}^{K}$, $\mathbf{X}_{i}=\{x_{i}^{k}\}_{k=1}^{K}$. $MP_{B}$ contains three parameters in the GPR model, i.e., the length-scale $l$, scale $\sigma^{2}_{f}$, and Gaussian noise $\sigma^{2}_{n}$. 

\begin{figure}[t!]
    \centering
    \subfigure[First scenario]{
    \includegraphics[width=.46\textwidth]{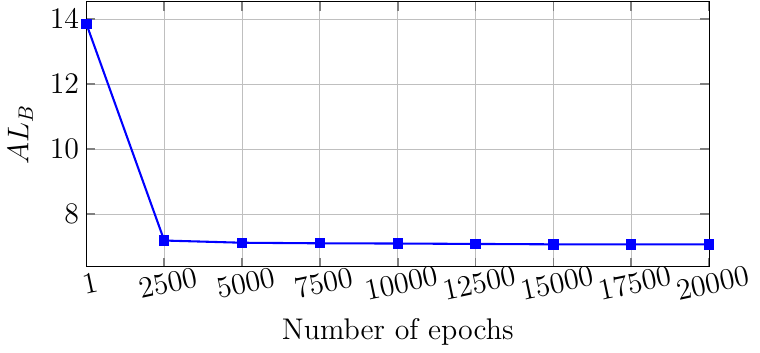}
    \includegraphics[width=.46\textwidth]{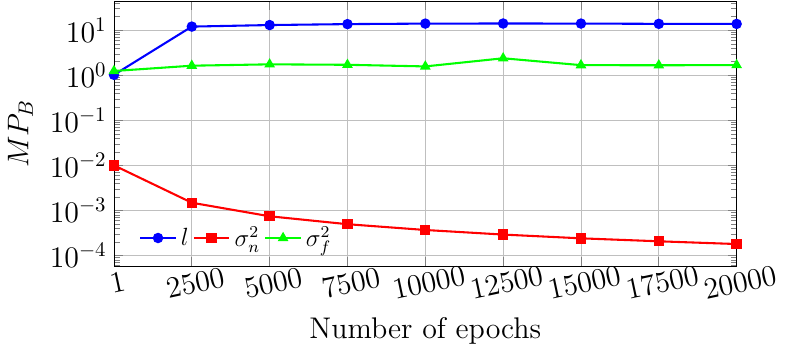}}
    \subfigure[Second scenario]{\includegraphics[width=.46\textwidth]{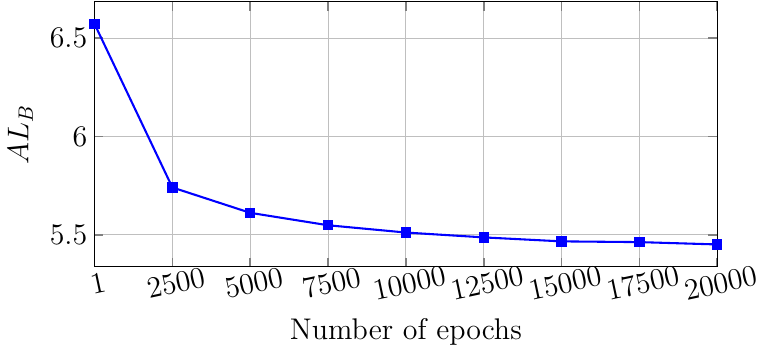}
    \includegraphics[width=.46\textwidth]{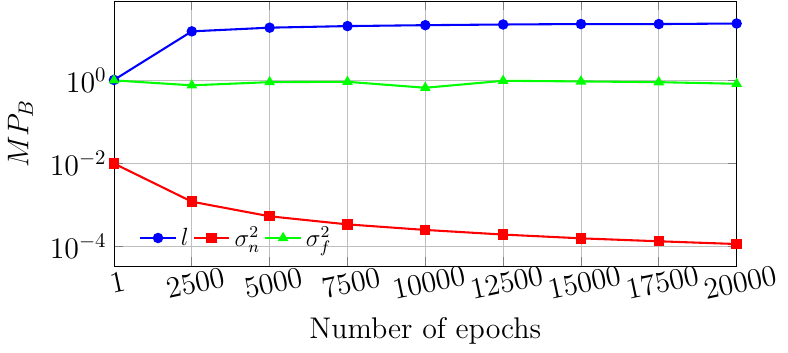}}
    \caption{Illustrative example of the $AL_{B}$ and $MP_{B}$ values during the \texttt{meta-learning} component in the two scenarios.}
    \label{fig:mpb-concern3-scenarios-ML}
\end{figure}

Let us consider an illustrative example, Fig.~\ref{fig:mpb-concern3-scenarios-ML} show the $AL_{B}$ and $MP_{B}$ during the \texttt{meta-learning} component obtained by \texttt{MLBO} in the two scenarios as shown in Fig.~\ref{fig:mpb-concern3-truef}, respectively. It is evident that the $AL_{B}$ gradually decreases until convergence is achieved at the end of the \texttt{meta-learning} component in both scenarios. We also observe that $AL_{B}$ in the second scenario requires more sampled batches to reach convergence than in the first scenario. Meanwhile, $MP_{B}$ is changing in a good direction for the new environment. Specifically, the length-scale increases gradually, the Gaussian noise decreases gradually, the scale does not change much, and they all level off at the end of the meta-learning. By using meta-learning, \texttt{MLBO} is able to learn initial model parameters for the new environment. This is accomplished concretely by optimizing the sum of loss across the query sets over related tasks, as described in~\pref{eq:loss} of~\pref{sec:meta-surrogate}. In contrast, the initial model parameters of the other peer algorithms are set as introduced in~\pref{sec:PSs}.  

\begin{figure*}[t!]
    \centering
    \subfigure[MLBO]{
    \includegraphics[width=5cm]{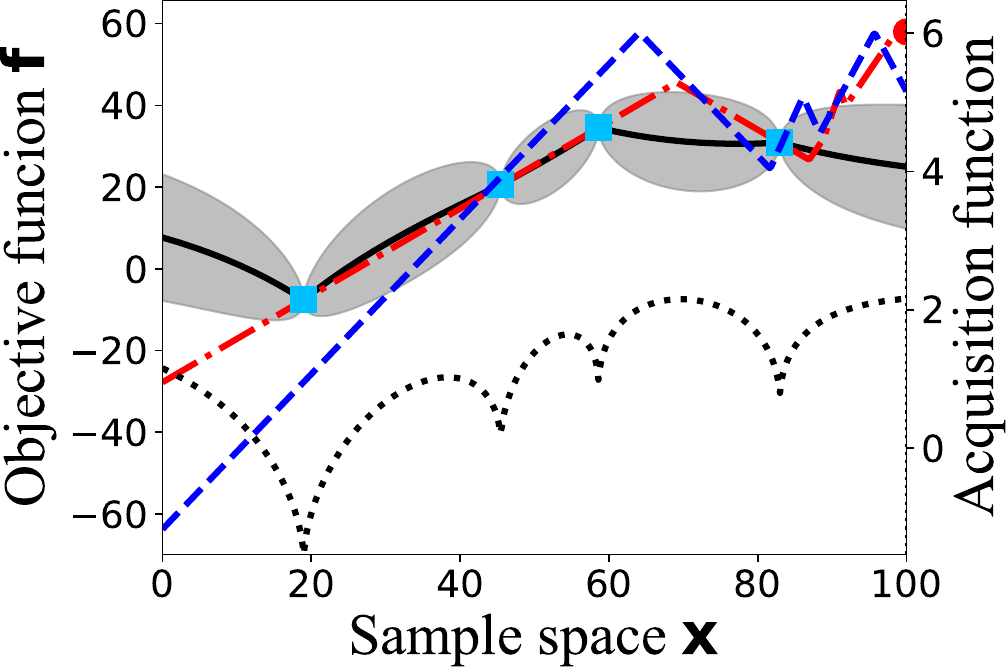}
    \includegraphics[width=5cm]{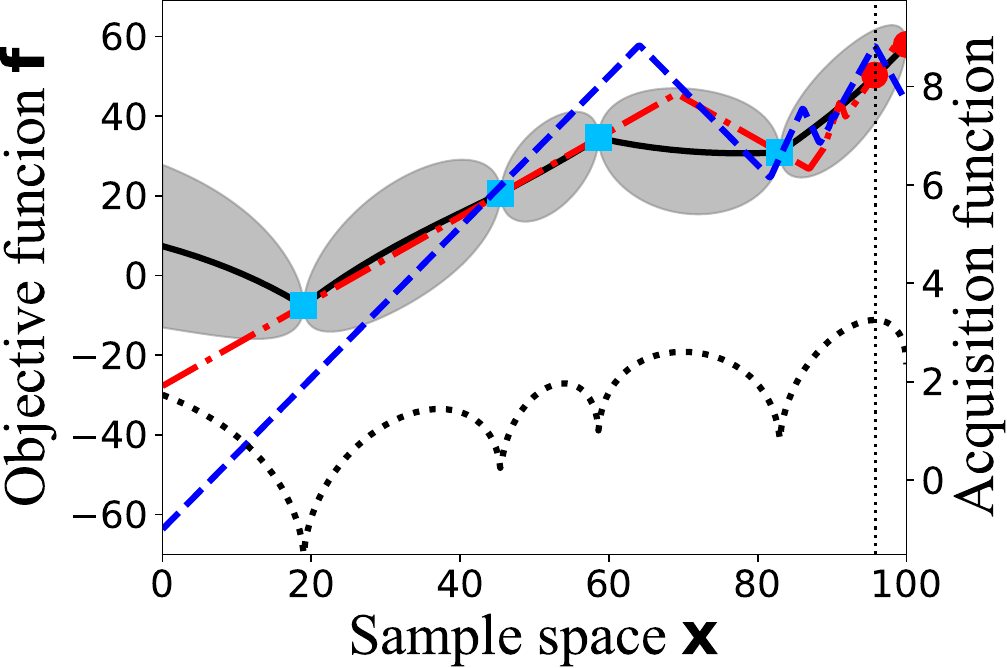}
    \includegraphics[width=5cm]{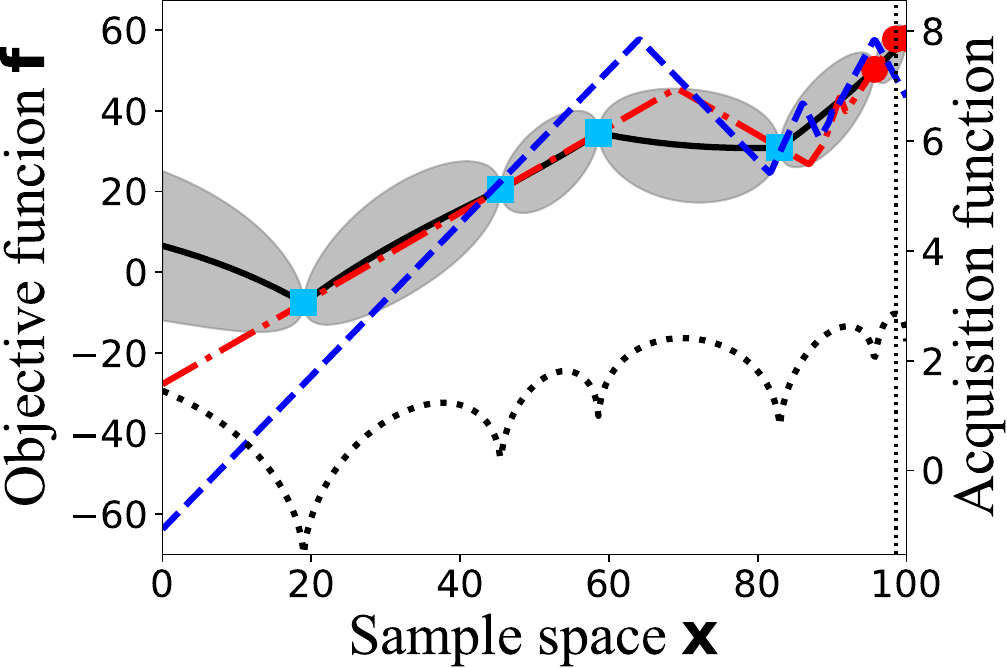}}
    \vfill
    \subfigure[RBO]{
    \includegraphics[width=5cm]{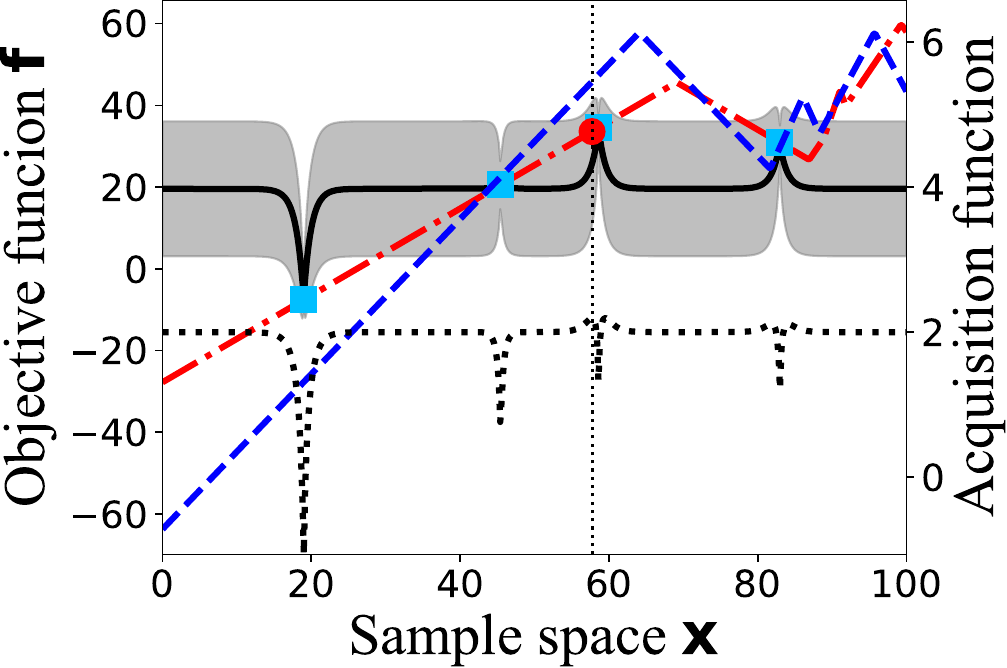}
    \includegraphics[width=5cm]{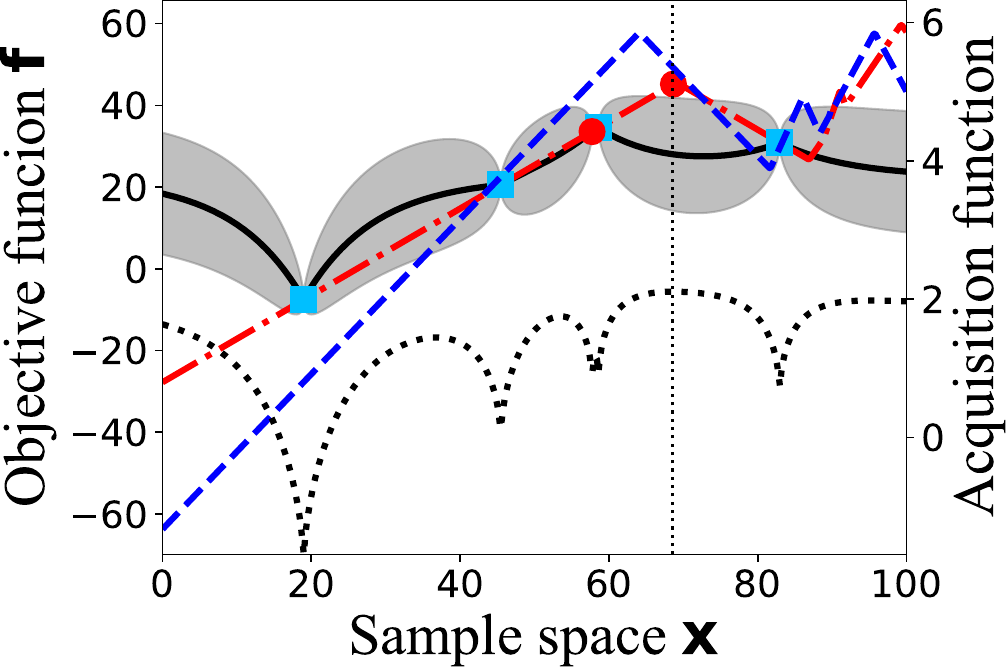}
    \includegraphics[width=5cm]{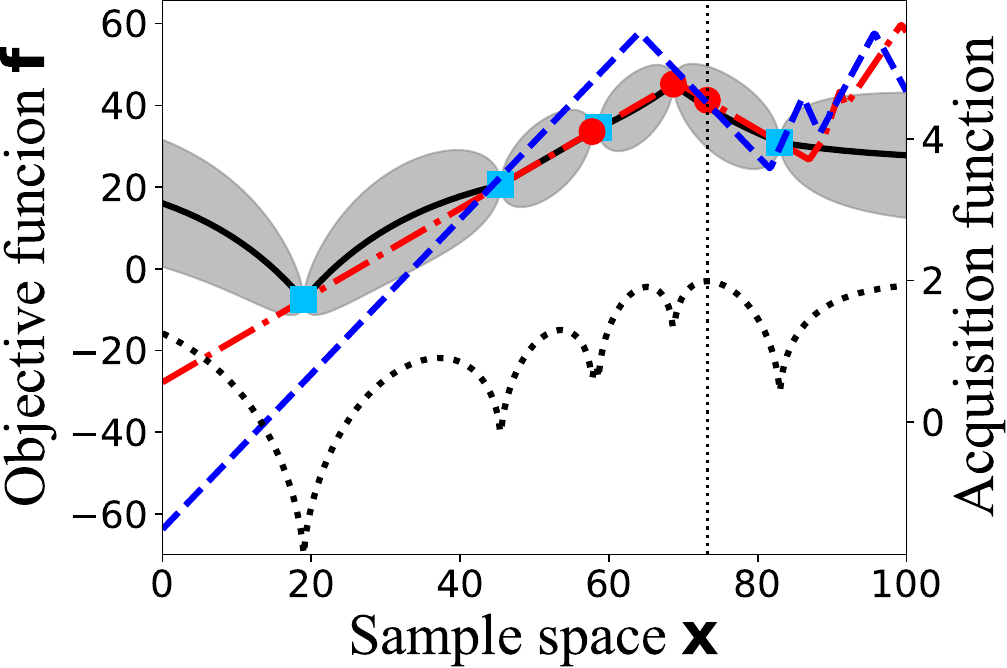}}
    \includegraphics[width=12.5cm]{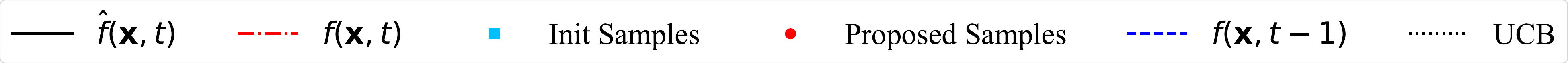}
    \caption{Illustrative example of the search dynamics of \texttt{MLBO} and \texttt{RBO} across three new samples after environmental changes in the second scenario.}
    \label{fig:mpb-concern3-scenarios2-MS}
\end{figure*}

\emph{2) Effectiveness of the adaptation component:} The above study has shown that the meta-learning helps the model learn initial parameters for the new environment. To verify the effectiveness of the learned initial parameters for subsequent \texttt{adaptation} component under the two scenarios shown in Fig.~\ref{fig:mpb-concern3-truef}, we look into the search dynamics of \texttt{MLBO} and \texttt{RBO} over three new samples through iteratively optimizing the UCB, respectively. As an example shown in Figs. 18 and Fig.~\ref{fig:mpb-concern3-scenarios2-MS}, \texttt{MLBO} can locate the new global optimum more quickly with the same number of sampling. Specifically, regardless of which scenario depicted in Fig.~\ref{fig:mpb-concern3-truef}, \texttt{MLBO} can locate near the global optimum at the first sampling. Then, \texttt{MLBO} probes quickly to the global optimum just at the third sample. This can be attributed to utilizing meta-learning to learn effective initial model parameters, enabling \texttt{MLBO} to enhance the quality of the surrogate model and achieve a jump start in the new environment for BO. In contrast, due to the manually given initial parameters, the model in \texttt{RBO} does not fit well in the initial stage, resulting in the first sampling being far from the global optimum. In addition, it is observed that the second scenario makes \texttt{RBO} more struggling during the search stage.

\begin{framed}
    \underline{Response to concern 3:} We derived the following two insights from the empirical study: 1) Parameter optimization of the surrogate model is a central pillar in the optimization procedure. A larger length-scale value and a smaller Gaussian noise value in a GPR model are helpful for fitting the GPR model. 2) The remarkable superior performance attained by the proposed framework can be attributed to the loss function in~\pref{eq:loss} as introduced in~\pref{sec:meta-surrogate}. It helps to learn effective initial model parameters for the surrogate in the new environment. Therefore, each of the proposed algorithm instances can achieve a jump start of optimization.
\end{framed}


\section{Conclusion}
\label{sec:conclusion} 
This paper has proposed a simple yet effective meta-learning-based optimization framework, called \texttt{MLO}, to solve expensive dynamic optimization problems. \texttt{MLO} encompasses three distinctive characteristics that set it apart from existing approaches. Firstly, it introduces a novel direction to address the challenges posed by solving expensive DOPs. Secondly, the proposed framework is flexible and allows any existing continuously differentiable surrogate model to be applied in a plug-and-play manner and further extended to data-driven evolutionary optimization or BO approaches. Lastly, by providing effective initial model parameters, \texttt{MLO} enables a jump-start of the optimization process in a new environment within limited computational resources. The experimental studies provide comprehensive evidences of the significant superiority of the proposed \texttt{MLO} agains state-of-the-art peer algorithms.


\bibliographystyle{IEEEtran}
\bibliography{IEEEabrv,casestudy}

\end{document}